\documentclass[10pt,journal,compsoc]{IEEEtran}
\usepackage{float}
\usepackage{graphics}
\usepackage{url}
\usepackage{graphicx}
\usepackage{multirow}
\usepackage{cite}
\usepackage[colorlinks=true, allcolors=blue]{hyperref}

\ifCLASSINFOpdf
 
\else
 
\fi

\begin{document}

\title{Learning to Predict Fitness for Duty using Near Infrared Periocular Iris Images}



\author{Juan Tapia,~\IEEEmembership{Member,~IEEE,}
        Daniel Benalcazar~\IEEEmembership{Member,~IEEE}, Andres Valenzuela, Leonardo Causa, Enrique Lopez Droguett~and~Christoph Busch,~\IEEEmembership{Senior Member,~IEEE}
\IEEEcompsocitemizethanks{\IEEEcompsocthanksitem Juan Tapia and Christoph Busch, da/sec-Biometrics and Internet Security Research Group, Hochschule Darmstadt, Germany.\\
E-mail: juan.tapia-farias@h-da.de \protect\\
\IEEEcompsocthanksitem Enrique Lopez Droguett, Department of Civil and Environmental Engineering, and Garrick Institute for the Risk Sciences, University of California, Los Angeles, USA.\protect\\
\IEEEcompsocthanksitem Daniel Benalcazar, Andres Valenzuela, Department of Mechanical Engineering, Universidad de Chile Av. Beauchef 851, Santiago, Chile.\protect\\
\IEEEcompsocthanksitem Leonardo Causa, TOC Biometrics, R\&D Center SR-226.
}%
\\
\textbf{**This work has been submitted to journal for possible publication. Copyright may be transferred without notice, after which this version may no longer be accessible.**}

\thanks{Manuscript received September 02, 2022; revised xxx, 2022.}}

\markboth{Journal of \LaTeX\ Class Files,~Vol.~14, No.~8, August~2015}%
{Tapia \MakeLowercase{\textit{et al.}}: Bare Demo of IEEEtran.cls for Computer Society Journals}

\IEEEtitleabstractindextext{%
\begin{abstract}
This research proposes a new database and method to detect the reduction of alertness conditions due to alcohol, drug consumption and sleepiness deprivation from Near-Infra-Red (NIR) periocular eye images. The study focuses on determining the effect of external factors on the Central Nervous System (CNS). The goal is to analyse how this impacts iris and pupil movement behaviours and if it is possible to classify these changes with a standard iris NIR capture device. This paper proposes a modified MobileNetV2 to classify iris NIR images taken from subjects under alcohol/drugs/sleepiness influences. The results show that the MobileNetV2-based classifier can detect the Unfit alertness condition from iris samples captured after alcohol and drug consumption robustly with a detection accuracy of $91.3\%$ and $99.1\%$, respectively. The sleepiness condition is the most challenging with $72.4\%$. For two-class grouped images belonging to the Fit/Unfit classes, the model obtained an accuracy of 94.0$\%$ and 84.0$\%$, respectively, using a smaller number of parameters than the standard Deep learning Network algorithm. This work is a step forward in biometric applications for developing an automatic system to classify "Fitness for Duty" and prevent accidents due to alcohol/drug consumption and sleepiness. 
\end{abstract}

}

\maketitle

\IEEEdisplaynontitleabstractindextext

%
\IEEEpeerreviewmaketitle

\IEEEraisesectionheading{\section{Introduction}\label{sec:introduction}}

\IEEEPARstart Iris recognition systems have been used mainly to recognize the cooperative subjects in controlled environments for access control to countries, buildings and offices using near-infra-red capture devices \cite{Jain2021,TapiaPerezBowyer2016}. With the improvements in iris recognition performance and reduction cost for iris acquisition devices, the technology will witness broader applications and may be confronted with new challenges \cite{Jain2021}. One such a kind of challenge is identifying if a subject is under alcohol, drug effects, or even under sleep deprivation and sleep restriction conditions. This area is known as "Fitness For Duty" \cite{FFD, murphy} and allows to determine whether the subject is physically able to perform his or her task \cite{murphy}. 

Fitness For Duty (FFD) \cite{FFD, murphy, Benderoth3-min} is a technique used in the context of the occupational test. These tests are applied to workers to describe a set of tools that help to evaluate a subject's condition considering physical, alertness, and emotional level which is required for a specific job. 
Determining if the person is "Fit" or "Unfit" to perform the job is a difficult task. Stay fit means completing the job's duties in a safe, secure, productive, and effective manner. In the state-of-the-art, several performance tests have been proposed as FFD, including psychomotor tasks, temperature sensor, EEG, finger tapping, pattern comparison, smart band wrist, and in-cab monitoring, among others \cite{ miller1996fit, serra2007criteria}. 
These systems deliver an answer based on the statistical behaviour threshold of "Fit" for the control subject and "Unfit" for alcohol/drug/sleepiness. It is essential to highlight that FFD does not have any relationship with the automatic alcohol test or drug test that measures alcohol/drug percentage in blood.

Many of these tests are reactive tasks and have demonstrated sensitivity to detect various job-related impairments, and also have shown some critical problems related to the type of device response and the possibility of impersonation. For instance, a wrist smart-band can be shared with another partner or removed to avoid identification. Therefore, including a biometric identification process in the FFD is crucial for the reliability and credibility of test results.

Today, primarily reactive devices are used, i.e., systems that act once the risk event has been generated, so it is necessary to have proactive solutions that allow measurements to be taken at the beginning of the working day. 

Regarding the impersonation, the systems cannot identify the user performing the test, and therefore it is possible to alter the results using a third person. At this point, biometrics plays a fundamental role in ensuring the person performing the FFD. Observing a biometric characteristic such as the iris may help to identify the worker without removing the personal protection devices.

The concept of biometrics is known as a set of intrinsic and behavioural characteristics that can be used to verify an individual's identity. All human beings have unique morphological characteristics that differentiate and identify us, the shape of the face, the geometry of parts of our body, like hands, our eyes, and perhaps the best known, the fingerprint \cite{Jain2021}. 

Biometric features used for recognition purposes can be extracted from images representing such morphological characteristics. For example, the face is a variable characteristic that depends on many factors, which can be easily changed by involuntary or voluntary action of the individual, simulating sad/happy faces or including COVID, dust, gas mask and others. 

To determine whether a worker is Fit to perform a task, it is necessary to observe that repetitive behaviours or biometric factors are manifested through the body, establishing the relationship between the cause and the effect of his/her behaviour. 

\begin{figure}[H]
\centering
\includegraphics[scale=0.4]{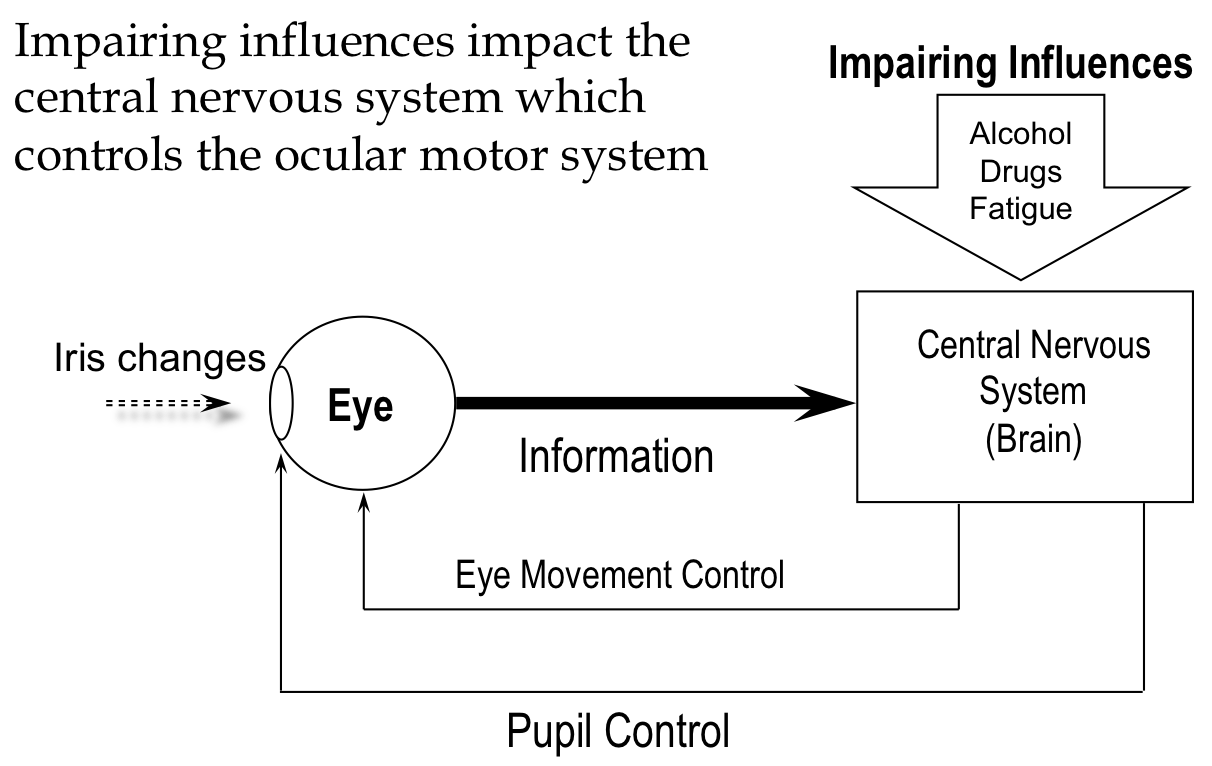}
\caption{\label{fig:CNS} Block diagram of the influence of the external factors in the Central Nervous System.}
\end{figure}

The iris and pupil movements are controlled by the Central Nervous Systems (CNS)\cite{Adler1985}. In this condition, the subject cannot control the pupil or iris movement. This action is initiated automatically by an external factor such as light or alcohol, drug consumption and others. See Figure \ref{fig:CNS}. Therefore, the iris is highly reliable for measuring the Fitness For Duty of the subject. This technique has been previously used but with optical equipment using a light beam to measure the changes of the iris, similar to the driver's license test \cite{SunYaoJi2012}. 

In order to create a framework to assess the worker’s FFD based on iris measurements, it is necessary to capture $N$ frames, detect both eyes, and segment the images to localise the pupil and the iris to measure the changes over time. Segmenting this kind of image is not a trivial task because the method needs to be efficient in the number of parameters to be implemented in a regular iris capture device. Most of these sensors are mobile devices self-integrated with a limited size of memory \cite{tapia2021semantic}.

\begin{figure}[]
\centerline 
{\includegraphics[scale=0.40]{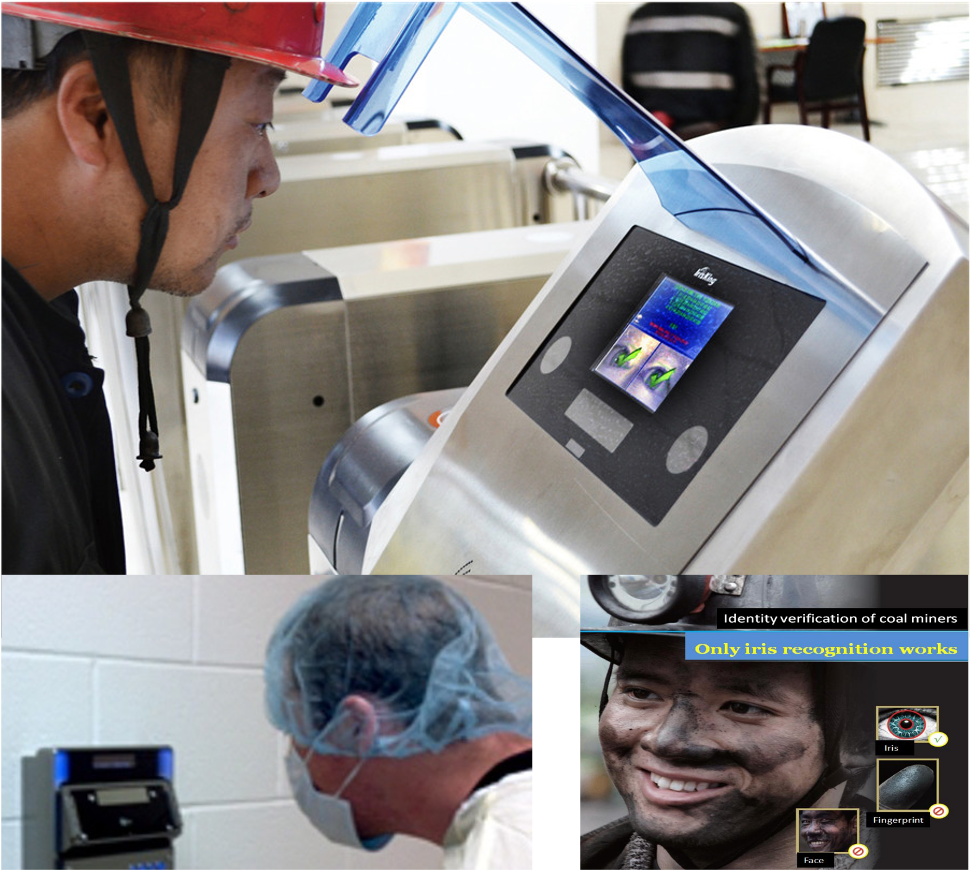}}
\caption{Example of iris recognition on industrial applications \cite{photo_iris}.}
\label{fig:chin-iris}
\end{figure}

A contactless capture device such as an iris recognition system is depicted in Figure \ref{fig:chin-iris}.

Recently, automatic pupil segmentation software is attracting the attention of many researchers to find the precise measurement of the pupil radii to apply to biometric analytics as part of medical, entertainment applications, and virtual reality lenses, among others \cite{Valenzuela}. In the biomedical field, there is a vital requirement to develop precise and automatic segmentation systems to capture saccade velocity, latency, the diameter of the iris, and pupil \cite{Czajka2015, Pinheiro2015alcoholvideo, AroraVatsaSinghEtAl2012}.

While alcohol's influence on iris recognition has not been extensively investigated, even less work has been done to analyse and counteract the effect of drug-induced pupil dilation, and sleepiness on iris images \cite{Tomeo-ReyesRossChandran2016}. 

Tomeo et al. \cite{Tomeo-ReyesRossChandran2016} focused on pupil and iris behaviour induced by the use of alcohol agents, but it could also be extended and allied to drug consumption as it has been shown that drug-induced pupil dilation affects iris recognition performance negatively. However, no solution has been investigated to counteract its effect on iris recognition. Note that alcohol/drugs may be used by an adversary to mask their identity from an iris recognition system \cite{AroraVatsaSinghEtAl2012}. These agents can be easily obtained online without a medical prescription. Hence, there is a need to understand and counteract the effect of alcohol and drugs on iris recognition.

An estimated 15-25\% of the workforce works in shifts \cite{BALASUBRAMANIAN202052, WICKWIRE20171156, Lukas}. Working in rotating shifts with work at night can be a significant risk factor, if the work duty is associated with critical infrastructure, equipment, and people. Also, several studies have shown an essential correlation between shiftwork, work at night, and alcohol and drug consumption as important triggers for occupational accidents \cite{Lukas}. Workplace alcohol use and impairment directly affect an estimated 15\% of the U.S. workforce; about 10.9\% work under the influence of alcohol or with a hangover \cite{Frone}. The Australian Government alcohol guidelines report shows 13\% of shift-workers, and 10\% of those on standard schedules reported consuming alcohol and drugs at levels risky for short-term harm \cite{DorrianandSkinner}.

This work proposes a new database and a modified multiclass convolutional neural network to classify control, alcohol, common drugs, and sleep deprivation iris samples. The main contributions of this work can be summarised as follows:

\begin{itemize}
    \item \emph{Overview}: 
    A comprehensive overview and literature survey of works that use alcohol, drug and sleepiness on Fitness For Duty and biometrics is presented.
    
    \item \emph{Architecture}:
    A multi-class MobileNetV2 architecture is proposed. This consists of a modified network trained from scratch, which is utilised to differentiate between Fit (control) and Unfit (alcohol, drug and sleepiness) subjects. New weights will be available.

    \item \emph{Experiments}: An extended set of Deep Neural Networks (DNN) experiments was conducted, selecting the best image using the sharpness metric and selecting three and five NIR images randomly and sequentially. Also, we evaluated the Maximum and Average score function for Unfit (alcohol, drugs, sleepiness) versus Fit.

    \item \emph{Weights}: Balanced class weights were used in order to correctly represent the number of images per class. Most of the databases are unbalanced according to the real application problem. Weighted classes help to balance the dataset and to get realistic results.

    \item \emph{Database}: A new dataset was captured for this paper to classify Fit and Unfit per image, meaning every subject has at least 90 images representing 5 seconds of capture. This database will be available to other researchers upon request for research purposes only (See Section~\ref{sec:database}).

    \item \emph{Data-Augmentation (DA)}: A three-level DA technique to train the modified MobileNetV2 networks was used. These images allow the network more challenging scenarios considering blurring, Gaussian noise, coarse occlusion, zoom, and others.

\end{itemize}

This paper aims to develop an efficient framework to classify single or multiple frames in subjects with alcohol, drug consumption, and sleepiness condition. The main objective is to estimate and extract the most relevant features from the iris and pupil to predict the Fitness For Duty for each person in order to save lives.

The remaining of this article is organised as follows: Section \ref{sec:related_work} describes state-of-the-art. The proposed method for FFD detection is presented in Section \ref{sec:method}. Section \ref{sec:database} explains the database. The experimental results are discussed in Section \ref{sec:result}. Conclusions and remarks are given in in Section \ref{sec:conclusion}

\begin{figure*}[]
\begin{centering}
\includegraphics[scale=0.5]{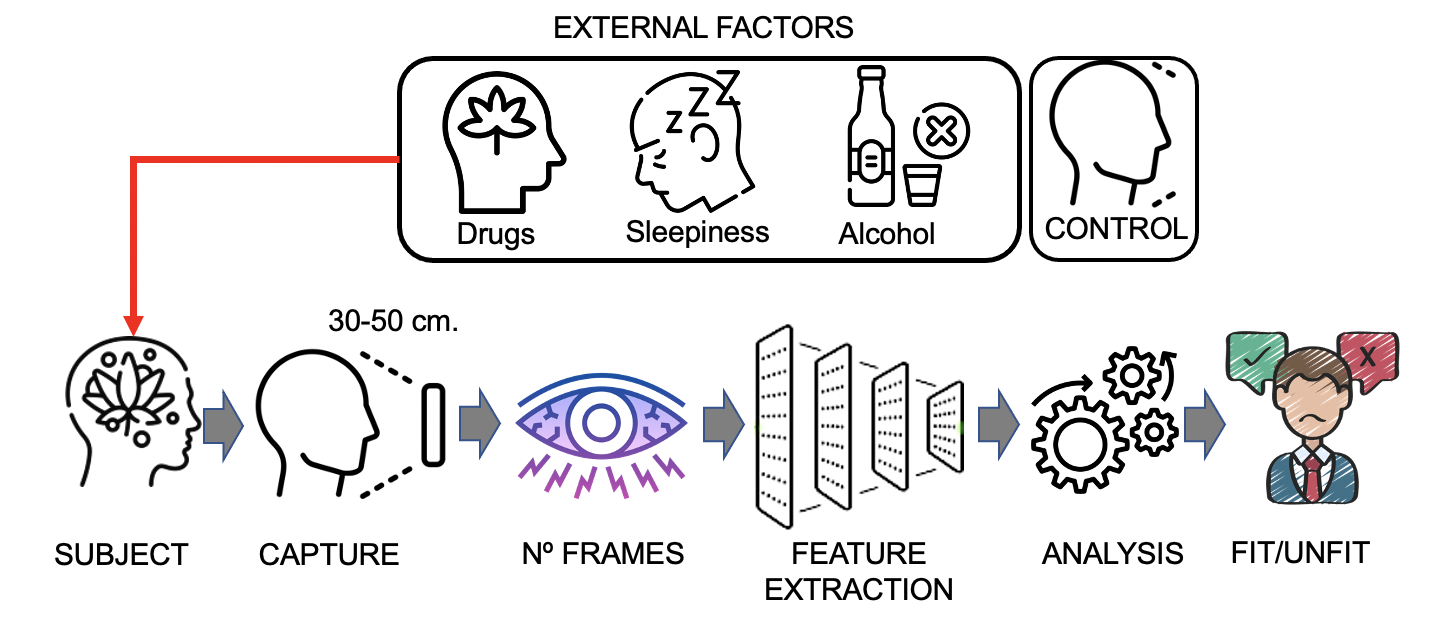}
\par\end{centering}
\caption{\label{framework1}Block diagram of the proposed FFD model based on periocular eyes images. The input is NIR image frames followed by the eyes detector, feature extraction, and the FFD model multiclass classifier. The system delivers the FFD level: the Fit/Unfit indicator.}
\end{figure*}

\section{Related work}
\label{sec:related_work}

This section reports a brief review of alcohol, drugs and sleepiness's influence on iris changes and their effects on "Fitness for Duties" are reported. 

Alcohol and drugs are Central Nervous System (CNS) depressants and stimulants, respectively. Alcohol diminishes environmental awareness, response to sensory stimulation, cognitive functioning, spontaneity, and physical activity. In high doses, alcohol can produce increasing drowsiness, lethargy, amnesia, anti-epileptic effects, hypnosis, and anaesthesia. It is, therefore, not surprising that most countries restrict people from driving and/or operating dangerous equipment under the influence of alcohol \footnote{\url{http://www.who.int/gho/alcohol/en/}}. 
Alcohol (ethyl alcohol or ethanol) is mainly used for recreational purposes and is the second most widely used drug after caffeine \footnote{\url{https://www.bgsu.edu/recwell/wellness-connection/alcohol-education/alcohol-metabolism.html}}.

Alcohol has been shown to affect pupil sizes. Brown et al. \cite{BrownAdamsHaegerstrom-PortnoyEtAl1977} were one of the first to study the changes in pupil sizes under the influence of marijuana and alcohol.

The influence of alcohol, in particular, was reported by Arora et al. \cite{AroraVatsaSinghEtAl2012}. They presented a preliminary study of the influence of alcohol on an iris recognition system. The experiments were performed on the 'Iris Under Alcohol Influence' database (No available). Results show that comparing pre and post-alcohol consumption images increases the overlap between mated and not-mated comparison score distributions by approximately 20\%. These results were obtained using a relatively small database (220 pre-alcohol and 220 post-alcohol images obtained from 55 subjects). The subjects consumed about 200 ml of alcohol (with a 42\% concentration level) in approximately 15 minutes, and the images were captured 15-20 minutes after alcohol consumption. Arora et al. \cite{AroraVatsaSinghEtAl2012} suggest that about one in five subjects under the influence of alcohol may be able to evade identification by iris recognition systems.  

Bernstein et al. \cite{BernsteinMendezSunEtAl2017} used spectrogram images of size $224\times224$ from non-audio wave-forms to identify the presence of alcohol with Convolutional Neural Networks (CNN) and wearable sensors. They used 80 training images (40 positives, 40 negatives) and 20 test images (10 positives, 10 negative) and achieved a test accuracy after adjusting the learning rate, number of iterations, and gradient descent algorithm, as well as the time window and colouration of the spectrograms, of 72\% (n=20, 5 trials).

Ma et al. \cite{MaTanWangEtAl2004} analyse the impact of changes in pupil size on iris recognition performance and describe an efficient algorithm for iris recognition by characterising key local variations. The basic idea was that local sharp variation points, denoting the appearing or vanishing of an important image structure, were utilised to represent the characteristics of the iris. The whole procedure of feature extraction included two steps: 1) a set of one-dimensional intensity signals is constructed to effectively characterise the most important information of the original two-dimensional image; 2) using a particular class of wavelets, a position sequence of local sharp variation points in such signals is recorded as features. Experimental results on 2,255 iris images show that the performance of the proposed method was encouraging and comparable to the best iris recognition algorithm found in the current literature.

Drugs, on the other hand, are substances that impact the central nervous system and can have depressants, stimulants and/or disturbing effects. 
At the cerebral level, drugs act on neurotransmitters, altering their correct functioning, and affecting behaviour, mood or perception. In addition, they are likely to create physical and/or psychological dependence. Some of the most important effects are: i) inhibition and attenuation of brain mechanisms that serve to maintain wakefulness and can produce different degrees of effects from relaxation, drowsiness, sedation to anaesthesia or coma (depressant drugs), ii) euphoria, increased alertness and motor activity and decreased subjective sensation of fatigue and hunger (stimulant drugs), and iii) distortion of aspects related to perception, emotional states and organisation of thought \cite{jain2021drug, ray2021side}.

Several studies have reported the effect of various licit and illicit drugs on the pupil \cite{WallaceB.PickworthCone.1989, WallaceB.PickworthFudala.1990, Donald_JasinskiJohnson, Cone.1990}. They have shown a strong effect on pupil diameter reduced by all major opiates, including heroin, morphine, and codeine. 

Rownbothan et al. \cite{M.C.RowbothamJones1987, M.C.RowbothamBenowitz.1984} have reported significant increases in pupil diameter for subjects given intravenous and oral cocaine. Wallace et al. \cite{WallaceB.PickworthFudala.1990} has reported similar increases for amphetamines. 

Tomeo-Reyes et al.,  \cite{Tomeo-ReyesRossChandran2016} analysed the impact of drugs on pupil dilation and proposed the use of a biomechanical nonlinear iris normalisation scheme along with key point-based feature matching for mitigating the impact of drug-induced pupil dilation on iris recognition. They also investigated the differences between drug-induced and light-induced pupil dilation on iris recognition performance. The authors reported an average ratio ${p}$ of pupil and iris between (0.265 $< $ ${p}$ $<$ 0.515). A drug-related work summary is presented in Table \ref{tab:drugs-list}. 

\begin{table}[H]
\centering
\scriptsize
\caption{Related Work on Drugs.}
\label{tab:drugs-list}
\begin{tabular}{|c|c|c|}
\hline
Factor                                                                                  & Effect                                                                                                                                  & References                                                                                                  \\ \hline
Alcohol                                                                                 & \begin{tabular}[c]{@{}c@{}}Decreased Saccadic\\ Velocity.\\  Increase in Latency in \\ Chronic Alcoholics and\\ Pupil size\end{tabular} & \cite{tapia2021semantic, Navarro7877181, GunillaHaegerstrom1977, AroraVatsaSinghEtAl2012, MonteiroPinheiro2015, Amodio8515064, GUDDHURJAYADEV2021}                                                      \\ \hline
Barbiturates                                                                            & \begin{tabular}[c]{@{}c@{}}Decrease in Pupil \\ diameter Size\end{tabular}                                                              & \cite{Adler1985, GallowayAmoakuGallowayEtAl2006}                                                            \\ \hline
Cocaine                                                                                 & \begin{tabular}[c]{@{}c@{}}Increased Pupil \\ diameter Size\end{tabular}                                                                & \cite{Adler1985,GallowayAmoakuGallowayEtAl2006, M.C.RowbothamJones1987,  M.C.RowbothamBenowitz.1984}        \\ \hline
\begin{tabular}[c]{@{}c@{}}Heroin, Morphine\\ Codeine and Other \\ Opiates\end{tabular} & \begin{tabular}[c]{@{}c@{}}Reduced Diameter, \\ Saccadic Velocity\\  and Amplitude\end{tabular}                                         & \cite{Adler1985,GallowayAmoakuGallowayEtAl2006, WallaceB.PickworthCone.1989, WallaceB.PickworthFudala.1990} \\ \hline
Marijuana                                                                               & \begin{tabular}[c]{@{}c@{}}Reduced Constriction \\ Amplitude\end{tabular}                                                               & \cite{Adler1985, Tomeo-ReyesRossChandran2016}                                                               \\ \hline
Sedatives                                                                               & \begin{tabular}[c]{@{}c@{}}Decreased Pupil \\ diameter Size\end{tabular}                                                                & \cite{Adler1985, GallowayAmoakuGallowayEtAl2006}                                                            \\ \hline
\begin{tabular}[c]{@{}c@{}}Muscular\\ Relaxants\end{tabular}                            & \begin{tabular}[c]{@{}c@{}}Increased Pupil \\ diameter Size\end{tabular}                                                                & \cite{Adler1985, GallowayAmoakuGallowayEtAl2006}                                                            \\ \hline
\begin{tabular}[c]{@{}c@{}}Fatigue\\ Sleep \\ deprivation\end{tabular}                  & \begin{tabular}[c]{@{}c@{}}Decreased Pupil \\ diameter Size\\  Decreased Saccadic \\ Velocity\end{tabular}                              & \cite{Akerstedt2000, BaldwinDC2004,  FranzenPL2008, FranzenBuysseDahlEtAl2009}                              \\ \hline
\end{tabular}
\end{table}

\begin{table*}[]
\caption{Related Work on for Fitness For Duty.}
\label{FFD_list}
\resizebox{\textwidth}{!}{%

\begin{tabular}{|c|c|c|c|c|}
\hline
\textbf{Authors}        & \textbf{Application} & \textbf{Methods}                                                                                                                                                                                            & \textbf{Impaired-States Detectetd}                                                  & \textbf{Results}                                                              \\ \hline
Bakker et al. (2021) \cite{9468428}    & Reactive device      & \begin{tabular}[c]{@{}c@{}}Feature extraction module combined with sleepiness detection module \\ based on Karolinska Sleepiness Scale (KSS)\end{tabular}                                                   & Fatigue/drowsiness                                                                  & 72\% accuracy                                                                 \\ \hline
Fox et al. (2015) \cite{fox2015407}      & Consultancy model    & \begin{tabular}[c]{@{}c@{}}Activity, Fatigue, Task Effectiveness (SAFTE) Model: a biomathematical\\ analysis based on Three Process Model using real-time data\end{tabular}                                 & Fatigue/drowsiness                                                                  & \begin{tabular}[c]{@{}c@{}}80\% correct \\ detections\end{tabular}            \\ \hline
Balkin et al. (2004) \cite{BalkinComparative}    & FFD                  & \begin{tabular}[c]{@{}c@{}}Brief vigilance and attention task (Psychomotor Vigilance Task, PVT) \\ to measure the reaction time\end{tabular}                                                                & Fatigue/drowsiness                                                                  & Not reported                                                                  \\ \hline
Benderoth et al. (2021) \cite{Benderoth3-min} & FFD                  & 3 minutes PVT administered on a portable handheld device                                                                                                                                                    & \begin{tabular}[c]{@{}c@{}}Alcohol and \\ Fatigue/drowsiness\end{tabular}           & Not reported                                                                  \\ \hline
Hu and Min (2018) \cite{hu2018automated}      & Reactive device      & \begin{tabular}[c]{@{}c@{}}Feature detection on EEG signals and Gradient Boosting Decision \\ Tree (GBDT) detection\end{tabular}                                                                            & Fatigue/drowsiness                                                                  & 94\% accuracy                                                                 \\ \hline
Dey et al. (2019) \cite{dey2019real}       & Reactive device      & Image processing for facial points extraction and SVM classifier                                                                                                                                            & Fatigue/drowsiness                                                                  & 94.8\% accuracy                                                               \\ \hline
Persson et al. (2021) \cite{9055218}   & Reactive device      & \begin{tabular}[c]{@{}c@{}}Feature extraction based on Heart Rate Variability (HRV) and KNN,\\ SVM, AdaBoost, and Random Forest for final classification\end{tabular}                                       & Fatigue/drowsiness                                                                  & 85\% accuracy                                                                 \\ \hline
Kim et al. (2021) \cite{Kim9427252}      & FFD                  & Electroencephalography (EEG)-based deep learning algorithm                                                                                                                                                  & \begin{tabular}[c]{@{}c@{}}Alcohol, \\ Fatigue/drowsiness\\ and Stress\end{tabular} & \begin{tabular}[c]{@{}c@{}}88.1–96.4\% \\ accuracy\end{tabular}               \\ \hline
Tanveer et al. (2019) \cite{8846024}  & Reactive device      & \begin{tabular}[c]{@{}c@{}}Deep-learning-based driver-drowsiness detection for brain-computer \\ interface (BCI) using functional \\ Near-InfraRed Spectroscopy (fNIRS)\end{tabular}                        & Fatigue/drowsiness                                                                  & 99.3\% accuracy                                                               \\ \hline
\end{tabular}%
}
\end{table*}

Regarding Fitness for Duties, some related work has been previously analysed. Most of the methods available in the state-of-the-art are based on pupilometry and Psychomotor Vigilance tasks. A summary of FFD-related work is presented in Table \ref{FFD_list}. 

Our previous work \cite{causa} studied the analysis of eye frame sequences. This method showed that developing a behavioural curve to estimate FFD is possible. However, this method requires the detection, segmentation of pupil and iris for each frame, and the feature extraction of more than 30 different measures, determined according to the state-of-the-art- Then, this kind of method demands many resources to be implemented in a mobile device. 
\vspace{-0.3cm}

\section{Proposed method}
\label{sec:method}

The proposed FFD model can be described as a cascade of modules, as shown in Figure \ref{framework1}. The capture process allows one to find the eyes and crop them to the right and left eyes (in periocular images). A feature extraction module (MobileNetV2) uses cropped periocular images to automatically extract and weight a feature vector representing the classifier's input. Finally, the FFD model generates the final classification: Fit/Unfit indicator and level for each state (control, alcohol, drug and sleepiness). 

\subsection{Capture}
\begin{figure}[H]
\centerline
{\includegraphics[scale=0.35]{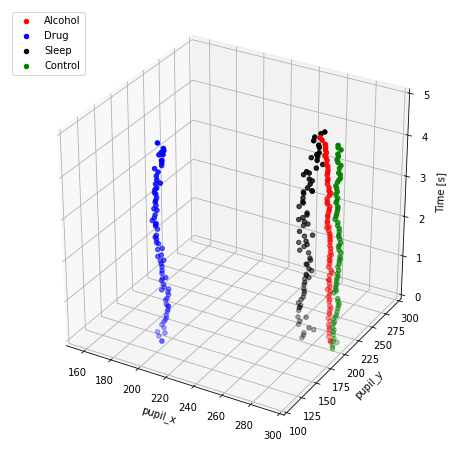}}
\caption{Movement representation for subjects under external factors in front of iris capture devices.}
\label{fig:disbalance}
\end{figure}

The iris capture process presents several challenges, such as the capture subjects reacting differently to the same quantity of alcohol and drugs. These changes do not allow the use of parametrical methods such as Osiris or commercial software \cite{tapia2021semantic}. Most of the time, people present semi-close eyes, which poses an extra difficulty. In the presence of alcohol and drugs, the volunteer in front of the sensor shows an involuntary disbalance. This adds blurring to the captured images, as shown in Figure \ref{fig:disbalance}.

\subsection{Image Pre-processing}

All the images in the database were pre-processed using a contrast-limited adaptive histogram equalisation algorithm (See sub-section \ref{clahe}) to improve the grey-scale intensity. During the training, a weighted factor was applied per class (See sub-section \ref{classweight}). Also, a higher number of filters was applied, using the MobileNetV2 alpha parameter, from the standard 1.0 up to 1.4. Both methods are leveraged to create a two-stage classifier that can detect Fit and Unfit scenarios. All the images were resized to 448$\times$448 for fine-tuning using new weights obtained from scratch according to the experiments. It is important to highlight that Imaginet pre-trained weights were used only when the model was explored using fine-tuning. Conversely, when the MobileNetV2 was trained from scratch, the Imaginet weights were not used. Preliminary results show a lower accuracy on traditional, ResNet34, ResNet50 and MobileNetv2.

\subsection{Contrast Limited Adaptive Histogram Equalisation (CLAHE)}
\label{clahe}

In order to improve the quality of the images and highlight texture-related features, the CLAHE algorithm was applied. This algorithm divides an input image into a M$\times$N grid. Afterwards, it applies equalisation to each cell in the grid, enhancing global contrast and definition in the output image. All the images were divided into 8$\times$8 sized cells.

\subsection{Class Weights}
\label{classweight}

A weight factor was estimated for each class according to the number of images of the class, helping to balance the database. Class weights are applied to the loss function, favouring under-representation and penalising over-representation of classes by re-scaling the gradient steps during training. See Equation~\ref{eq1}.

\begin{equation}
\label{eq1}
    Weight_i = \frac{Nsamples}{Nclasses \times samples_i}
\end{equation}

Where $Weight_i$ is the weight for class $i$, $Nsamples$ is the total number of images in the database, $Nclasses$ is the total number of classes in the database, and $samples_i$ is the number of samples of class $i$. The weight values associated with each class are the following: 

\begin{itemize}
    \item Class 0, Alcohol: 0.6474
    \item Class 1, Control: 0.7342
    \item Class 2, Drug: 1.820
    \item Class 3, Sleepiness: 1.838
\end{itemize}


\section{Database}

\label{sec:database}

One of this research's main challenges was developing the new database. This database contains images of alcohol, drug, and sleepiness. 
This task was a very demanding effort because all the participants were volunteers. The recruitment process was exhaustive and challenging, taking more than 1 year. For this research, we developed a new database called the "FFD NIR iris images Stream database" (FFD-NIR-Stream), containing 10-second stream sequences of periocular NIR images. The protocol was analysed and approved by the ethical committee of the Universidad de Chile.

\begin{figure}[htbp]
\centerline
{\includegraphics[scale=0.30]{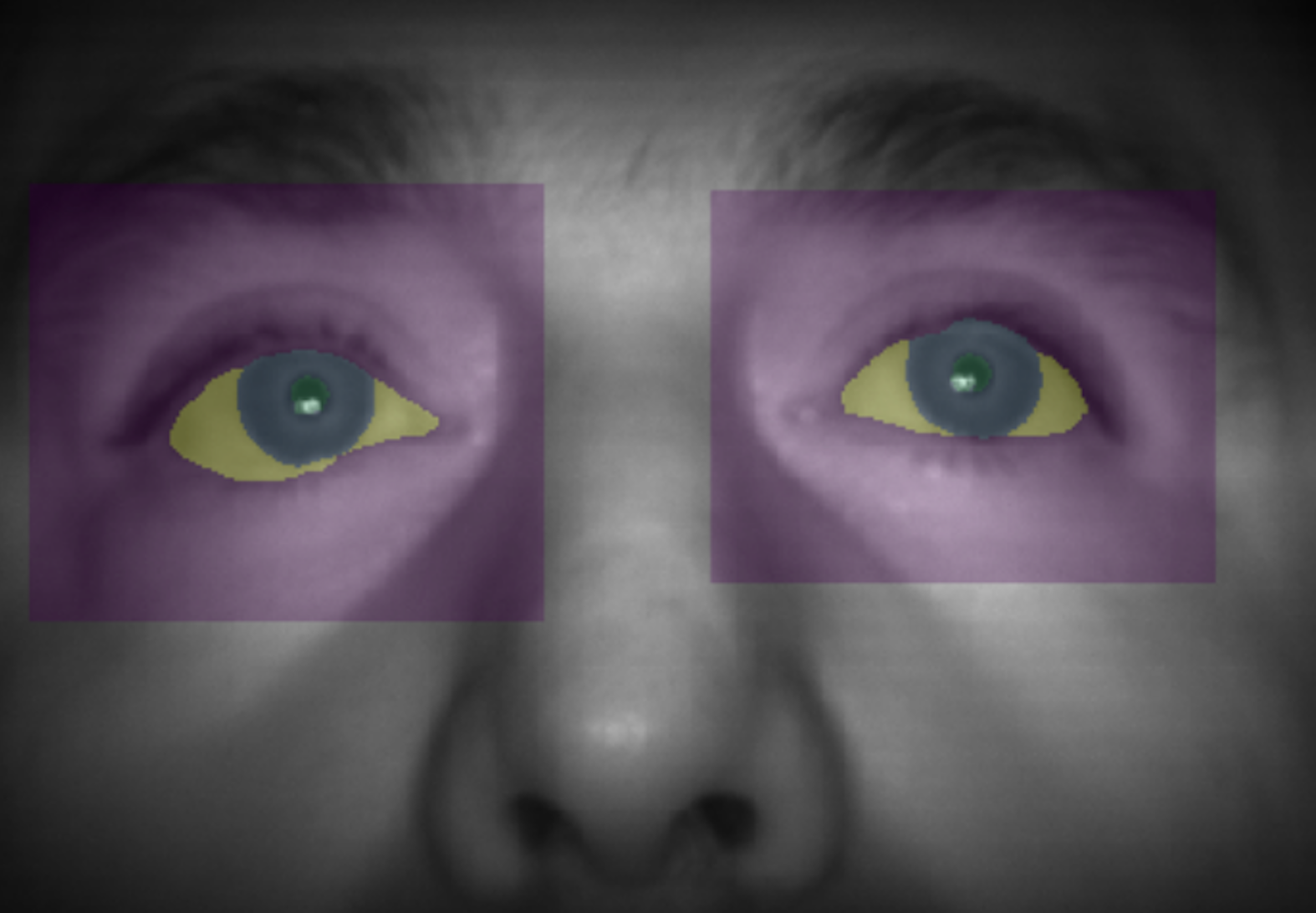}}
\caption{Example of a labelled periocular NIR image. The image shows both eyes and the corresponding labels to the right and left eyes.}
\label{capture}
\end{figure}

The binocular NIR image frames captured correspond to the periocular area (eye mask) to represent both eyes, pupils, iris, and sclera (see Figure~\ref{capture}).

Other images were acquired using a different capture device to get the area corresponding to each eye separately. For the acquisition of the image frames, the subjects were positioned in front of the capture device; the equipment detects the eyes and starts the recording.


This data makes it possible to perform the necessary processing to determine the iris and pupil behavioural parameters to be used in developing the models in this work.

The image sequences were captured by using four different devices\footnote{\url{https://www.iritech.com}}: i) Iritech MK2120UL (monocular), ii) iCAM TD-100A, iii) Iritech Gemini, and iv) Iritech Gemini-Venus. The room temperature and lighting (200 lux) were kept constant in the capturing process.


Four NIR image sequences in different conditions were registered:
\begin{itemize}
   \item 
Control DB: healthy subjects that are not under alcohol and/or drug influence and in normal sleeping conditions.
   \item 
Alcohol DB: subjects who have consumed alcohol or are in an inebriation state.
   \item 
Drugs DB: subjects who consumed some drugs (mainly marijuana) or psychotropic drugs (by medical prescription).    
   \item 
Sleep DB: subjects with sleep deprivation, resulting in fatigue and/or drowsiness due to sleep disorders related to occupational factors (shift structures with high turnover).
\end{itemize}

\subsection{Alcohol}
In the case of the alcohol database, the subjects were confronted with the following protocol: 
\begin{enumerate}
    \item The first NIR image sequence acquisition was made at time 0 (previous to alcohol consumption).
    \item All the Volunteers drank 200 ml of alcohol is up to 15 minutes. 
    \item The second acquisition was performed immediately after the alcohol intake was finished, i.e., 15 minutes after 0.
    \item The third acquisition was made 30 minutes after time 0.
    \item Fourth acquisition was made 45 minutes after time 0. 
    \item Finally, the fifth acquisition was made 60 minutes after time 0. 
\end{enumerate}
Thus, there were five sequences recorded of images of the subject under the effects of alcohol and one sequence of control images.

\subsection{Drugs}

\begin{figure}[htbp]
\centerline
{\includegraphics[scale=0.24]{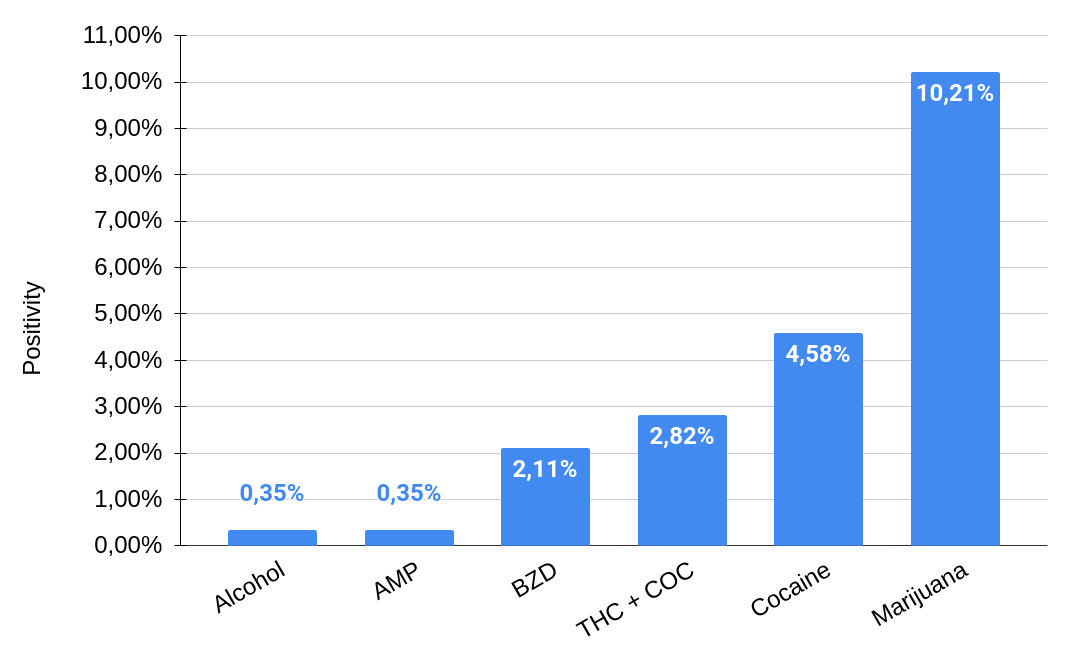}}
\caption{Percentage of positive for different kinds of the drug on the Chilean industry 2021. Based on \cite{drug-free}. AMP represent Amphetamine. BZD: Benzodiazepim.}
\label{img_drugs}
\end{figure}

According to the World Drug Report, 2021, of the United Nations Office on Drugs and Crime, cannabis\footnote{\url{https://www.unodc.org/unodc/en/data-and-analysis/wdr2021.html}} is the most widely consumed crop worldwide with an annual prevalence of 15\%, followed by pharmaceutical opioids and tranquillisers with a 5\% and 2.5\% of yearly prevalence, respectively. For this reason, about 95\% of the records in our database correspond to cannabis consumption. In contrast, the remaining 5\% corresponds to tranquillisers and more complex drugs (heroin and ecstasy). 
Drug users participated in the study as volunteers for the drug database acquisitions, and the image recordings took place at least 30 minutes after the initial consumption. This database focused on the most consumed drugs in the Chilean workplace according to the company drug-free \footnote{\url{https://drugfreeworkplace.cl/}}. See Figure \ref{img_drugs}.

\subsubsection{Sleep}

A particular image acquisition protocol for the sleep database was defined, in which tests were performed under controlled sleep deprivation conditions. These recordings were obtained on a specific group of subjects submitted to different sleep deprivation levels to evaluate the level of fatigue/drowsiness at different time intervals. The volunteers were monitored by using a smart band to measure the quantity and quality of sleep.
Subjects were grouped as follows:
\begin{enumerate}
    \item Total sleep deprivation
    \item Less than 3 hours of night sleep
    \item Between 3 and 6 hours of night sleep
    \item More than 6 hours of sleep (normal sleep).
\end{enumerate}

During the recording season, volunteers performed three daily image acquisitions: i) beginning of the working day, ii) post-lunch, and iii) at the end of the working day.

The FFD-NIR-Stream database consists of 1,510 eye-disjoint images (with 144,011 images). On average, 150 images were captured per subject. This process took ten seconds. The image sequences were divided into training, validation and test. In the case of the test set, the aim was to represent the actual proportion of Unfit instances, close to 15\%.

\begin{figure}[htbp]
\centerline
{\includegraphics[scale=0.40]{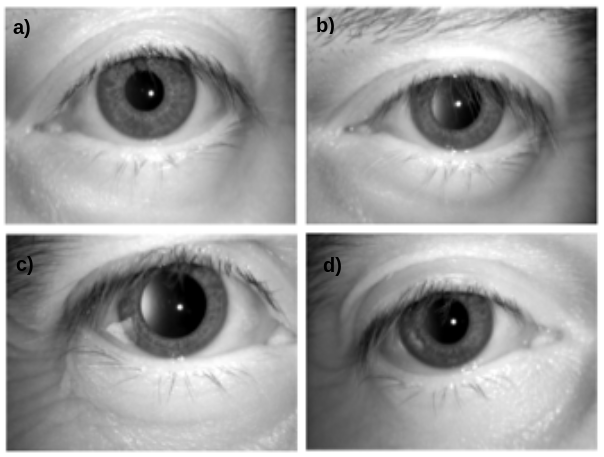}}
\caption{Examples of the NIR images captured. a) Control, b) Alcohol, c) Drug and d) Sleep images.}
\label{img_examples}
\end{figure}

Table~\ref{tab2} shows a summary of the total images, organised by conditions.  Figure~\ref{img_examples} shows examples of the acquired images for each of the conditions defined in the database.

\begin{table}[htbp]
\centering
\caption{Description of total NIR images recording by condition.}
\begin{tabular}{l|c|c|c}
\hline
\textbf{Conditions} & \textbf{Train set} & \textbf{Validation set} & \textbf{Testing set} \\ \hline
\hline
\textbf{Control} & 21,449 & 3,136 & 60,222 \\ \hline
\textbf{Alcohol} & 24,325 & 3,394 & 6,998 \\ \hline
\textbf{Drug} & 8,653 & 1,253 & 2,338 \\ \hline
\textbf{Sleep} & 8,568 & 1,140 & 2,535 \\ \hline
\textbf{Total} & 62,995 & 8,923 & 72,093 \\ \hline
\end{tabular}
\label{tab2}
\end{table}

This database was used to train and validate the different stages of the FFD model and classification stages.

\section{Metrics}
\label{sec:metric}

The False Positive Rate (FPR) and False Negative Rate (FNR) were reported as Error Type I and Error Type II. These metrics effectively measure to what degree the algorithm confuses presentations of Fit and Unfit images with alcohol, Drugs and Sleepness. The FPR and FNR are dependent on a decision threshold.

A Detection Error Trade-off (DET) curve is also reported for all the experiments. In the DET curve, the Equal Error Rate (EER) value represents the trade-off when the FPR and FNR. Values in this curve are presented as percentages. Additionally, two different operational points are reported. FNR\textsubscript{10} which corresponds to the FPR is fixed at 10\% and FNR\textsubscript{20} which is when the FPR is fixed at 5\%. FNR\textsubscript{10} and FNR\textsubscript{20} are independent of decision thresholds.

\section{Experiments and Results}
\label{sec:result}

\subsection{Experiment 1 - Scratch and Fine-tuning}

A modified MobileNetv2 network was used, trained from scratch and fine-tuning techniques. Several tests were performed, sequentially freezing an additional MobileNetV2 block in each one, from the bottom of the network to the top. For this experiment, the images were grouped into four classes: Control, Alcohol, Drug and Sleepiness. 

\begin{figure}[]
\centering
\includegraphics[scale=0.30]{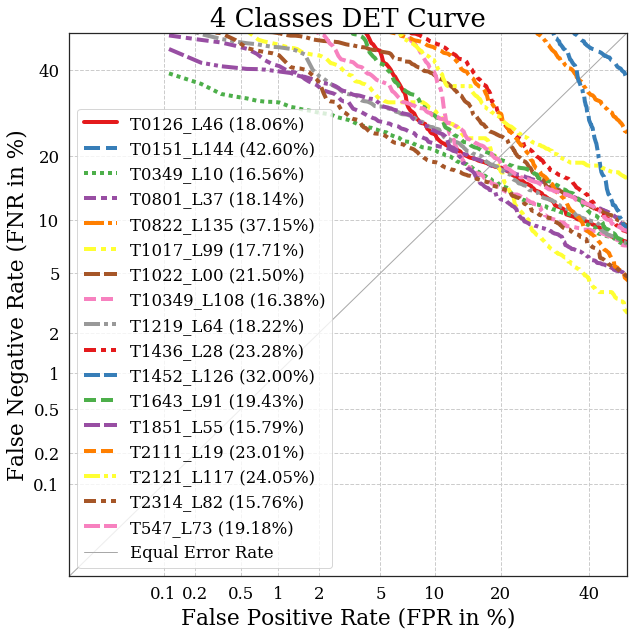}
\caption{\label{fig:DET_FT} DET curve for MobileNetV2 fine-tuning. All the layers were explored. In parenthesis, EER is shown in percentages. L{XX}: represents Layer number. L00 represent MobileNetV2 with all the layers frozen.}
\end{figure}

Figure \ref{fig:DET_FT} shows the DET curve for each layer. For fine-tuning the following layers were explored using $224\times224\times3$ image size: L00, 10, 19, 28, 37, 46, 55, 64, 73, 82, 91, 108, 117, 126, 135 and 144. Layer 82 reached the best results with an EER of 15.76\%. All the models use categorical cross-entropy and a learning rate of $1e-5$. It is essential to highlight that the best learning rate was obtained with a grid search method. The alpha value was set up to 1.0. 

From scratch, all the images were resized to $448\times448\times3$. The models use categorical cross-entropy and a learning rate of $1e-5$. The Average pooling was removed, and the alpha value was set up to 1.4. Thus, a bigger number of filters were used for feature extraction. 
For this case, all the images were considered new examples. No time correlation was used.

\begin{figure*}[]
\centering
\includegraphics[scale=0.26]{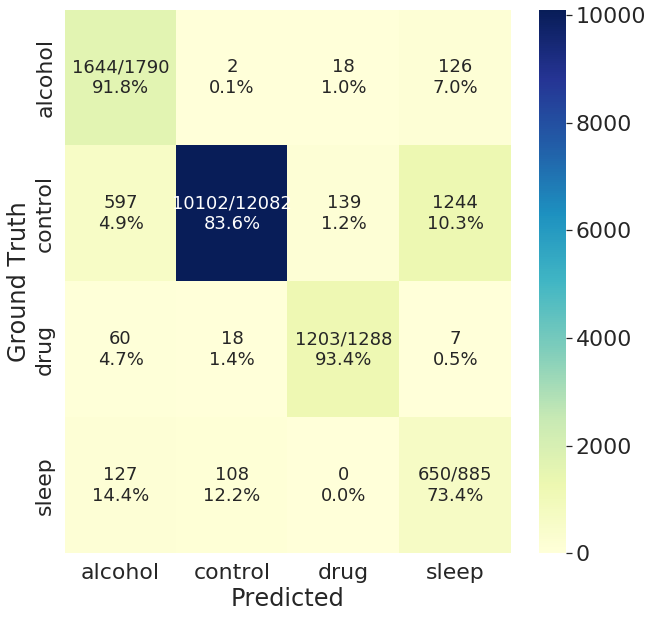}
\includegraphics[scale=0.26]{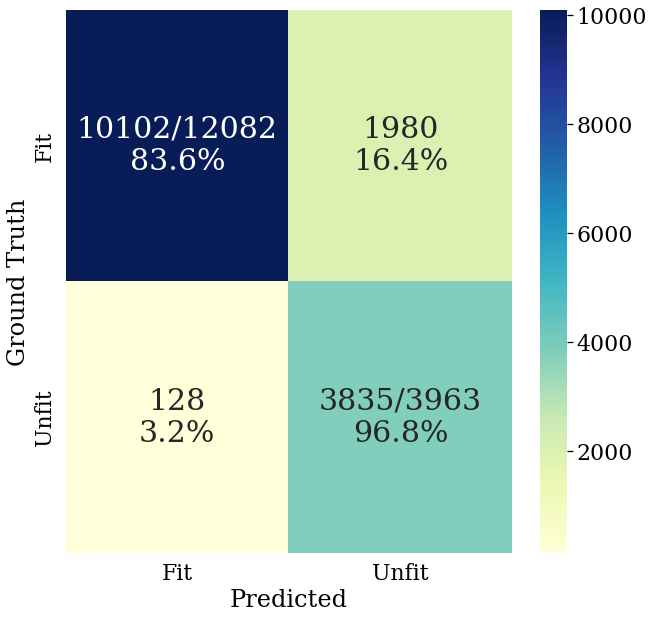}
\includegraphics[scale=0.26]{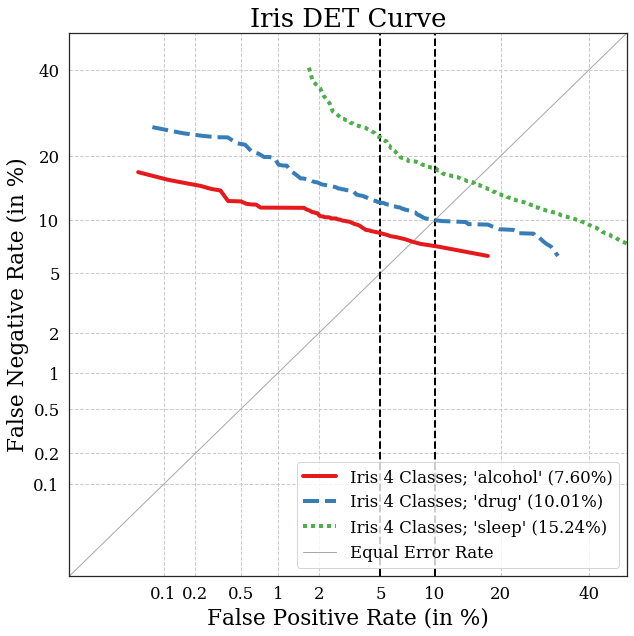}
\caption{\label{fig:MC_T0619} Confusion matrix and DET curves for model T0619 trained from scratch. Vertical black dot lines represent two operational points. FNR\textsubscript{10} and FNR\textsubscript{20}.}
\end{figure*}

Figure~\ref{fig:MC_T0619} shows the best scratch result model for the four class scenarios: control versus alcohol, drugs and sleepiness. Likewise, two confusion matrices, one showing four classes and the other grouping all Unfit condition classes, are presented. A Detection Error Trade-off (DET) curve is also shown. The best result for this experiment reaches an EER of only 7.60\% (red curve - alcohol), 10.01\% (blue curve - drug) and 15.24\% (green curve - sleep). An FNR\textsubscript{10} of 17.69\%, and an FNR\textsubscript{20} of 23.66\%. The best model uses an alpha value of 1.4, an initial learning rate of $1\times 10^{-5}$, and the Adam optimisation algorithm.

As the scratch model reached a better result than fine-tuning, an additional evaluation of 1, 3 and 5 frames was performed in experiments 2, 3 and 4. According to the results, using one image is enough to determine if one subject is Fit/Unfit. However, deciding which images could be used is necessary, i.e. select the first image or the images with the best sharpness, a random image, or a sequence of 3 or 5 frames. 

\subsection{Experiment 2 - Best frame}

According to the ISO/IEC 29794-6 standard, sharpness measures the degree of focus present in the iris image. Sharpness is calculated as a power spectrum function after filtering with a Laplacian of Gaussian operator for image quality according to the ISO/IEC 29794-6 standard. Therefore, such images may be used for biometric applications. For this experiment, all the captured images were evaluated online, and only the images with the best sharpness were saved for each capture session.

Figure \ref{fig:DET_no_sorted}, First row, shows the DET curves when sharpness metrics were used to select the best frame with the MAX function score. This means the final score corresponds to Max score among control, alcohol, drug and sleepiness.

Figure \ref{fig:DET_no_sorted}, Second row, shows the DET curves when sharpness metrics were used to select the best frame with the Average function. This means the final score corresponds to the average score among alcohol, drug, and sleepiness compared with the control. A Detection Error Trade-off (DET) curve is also shown. The EER from alcohol, drug and sleepiness is described in percentages. 

The Max function scores reached best results than Average scores with 3 and 5 images from the same subject.
The best result for this experiment reaches an EER of 18.80\% (red curve - alcohol), 14.29\% (blue curve - drug) and 20.00\% (green curve - sleep). An FNR\textsubscript{10} of 28.13\%, and an FNR\textsubscript{20} of 36.29\%. 

\subsection{Experiment 3 - Random frame}

A modified MobileNetv2 multi-class network was used and trained from scratch. Several tests were performed, exploring different image sizes. For this experiment, the images were grouped into four classes: control, alcohol, drug and sleepiness. The best frame was selected randomly for each subject.

Figure \ref{fig:DET_no_sorted} the third row shows the DET curves when a random selection was applied to select 1 (the first), 3 and 5 frames per subject using a MAX function. 

Figure \ref{fig:DET_no_sorted}, fourth row, shows the DET curves when a random selection was applied to select 1 (the first), 3 and 5 frames per subject using an Average function. The EER from alcohol, drug and sleepiness is described in percentages.

For the random frame, the best result for this experiment reaches an EER of 20.39\% (red curve - alcohol), 18.75\% (blue curve - drug) and 26.32\% (green curve - sleep). An FNR\textsubscript{10} of 34.64\%, and an FNR\textsubscript{20} of 34.69\%.

\subsection{Experiment 4 - Sequential frames}

For sequential frames, the images were sorted according to the captured frame time. In total, the 797 captured sequences were divided into train, test and validation. These datasets are disjoint with respect to the involved subject. This means that all images belonging to one subject are used in only one set. This analysis makes sense when time is considered a new variable. However, this kind of analysis is demanding because it is required to detect the eyes, segment, and calculate the iris size for each frame. Then, this is not possible for a mobile device. An exhaustive analysis was made in our previous works \cite{tapia2021semantic, causa}.

In order to evaluate the small sequences without time correlation, only the first three and five frames are used to compute an Average and Max classification.

\begin{figure*}[]
\centering

\includegraphics[scale=0.35]{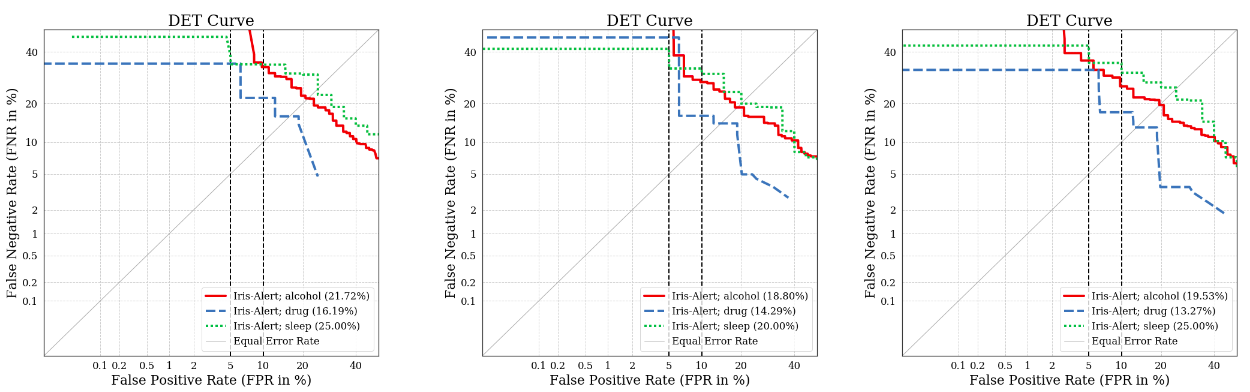}
\includegraphics[scale=0.35]{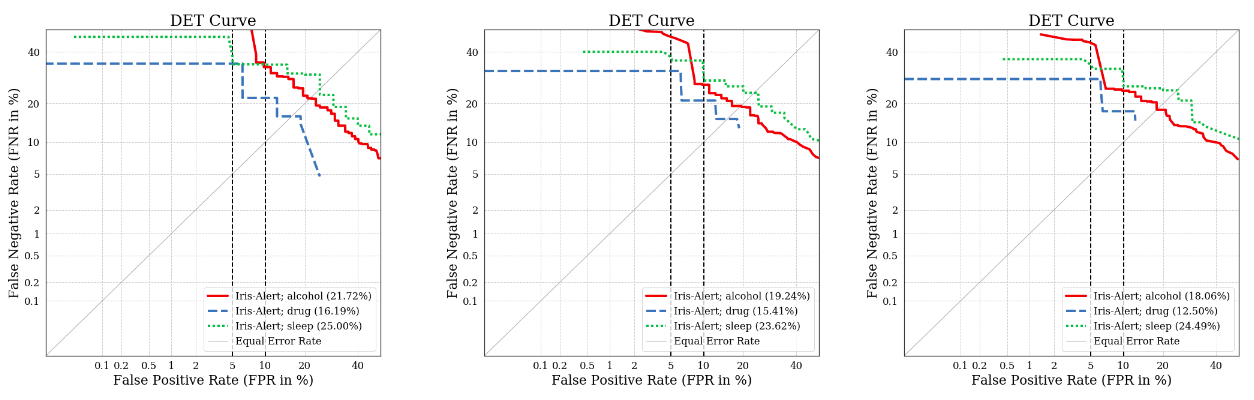}
\includegraphics[scale=0.35]{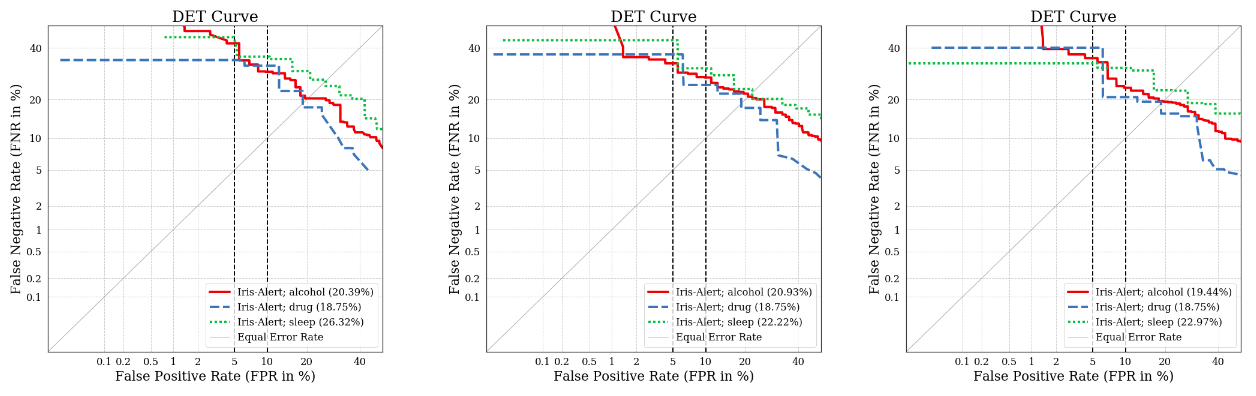}
\includegraphics[scale=0.35]{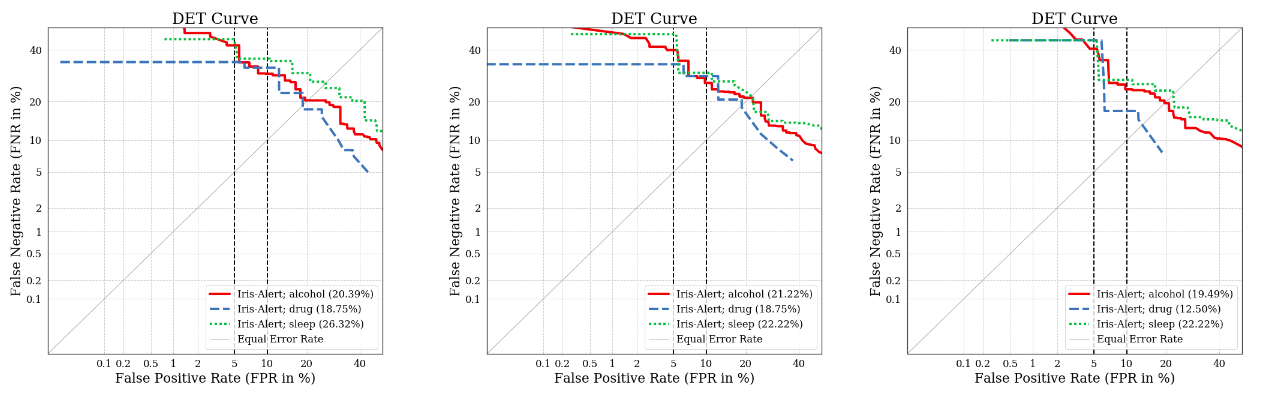}

\caption{\label{fig:DET_no_sorted} Summary EER results. The first row shows DET curves from 1 (best frame), 3 and 5 frames were organised by sharpness and Max operation was applied for the results.
Second-row shows DET curves from 1, 3 and 5 frames were organised by the sharpness and Average operation for the classification results. Third-row shows DET curves from randomly selected 1, 3 and 5 frames and Max operation was applied for the results.
Fourth-row shows DET curves from randomly selected 1, 3 and 5 frames and Average operation was applied for the classification results. The EER is shown in parenthesis for each curve.}
\end{figure*}

Figure \ref{fig:DET_no_sorted}, First row, shows the DET curves when sharpness metrics were used to select the 1, 3 and 5 best frames with the MAX function. 

Figure \ref{fig:DET_no_sorted}, Third row, shows the DET curves when sharpness metrics were used to select the 1, 3 and 5 best frames with the Average function. The EER from alcohol, drug and sleepiness is described in percentages. 

The Average function scores reached best results than the Average scores with 3 and 5 images from the same subject.
The best result for this experiment reaches an EER of 19.49\% (red curve - alcohol), 12.50\% (blue curve - drug) and 22.22\% (green curve - sleep). An FNR\textsubscript{10} of 31.19\%, and an FNR\textsubscript{20} of 35.27\%. 
Overall, the model trained from scratch (Experiment 1) reached
best results than Sharpness, Random and small Sequences (Experiments 2, 3 and 4).

\subsection{Visualisation}

A t-SNE map was used to visualise how the data is projected to a 2D plot. This method shows non-linear connections in the data. The t-SNE algorithm calculates a similarity measure between pairs of instances in high dimensional and low dimensional spaces. It then tries to optimise these two similarity measures using a cost function.

\begin{figure}[]
\centerline
{\includegraphics[scale=0.20]{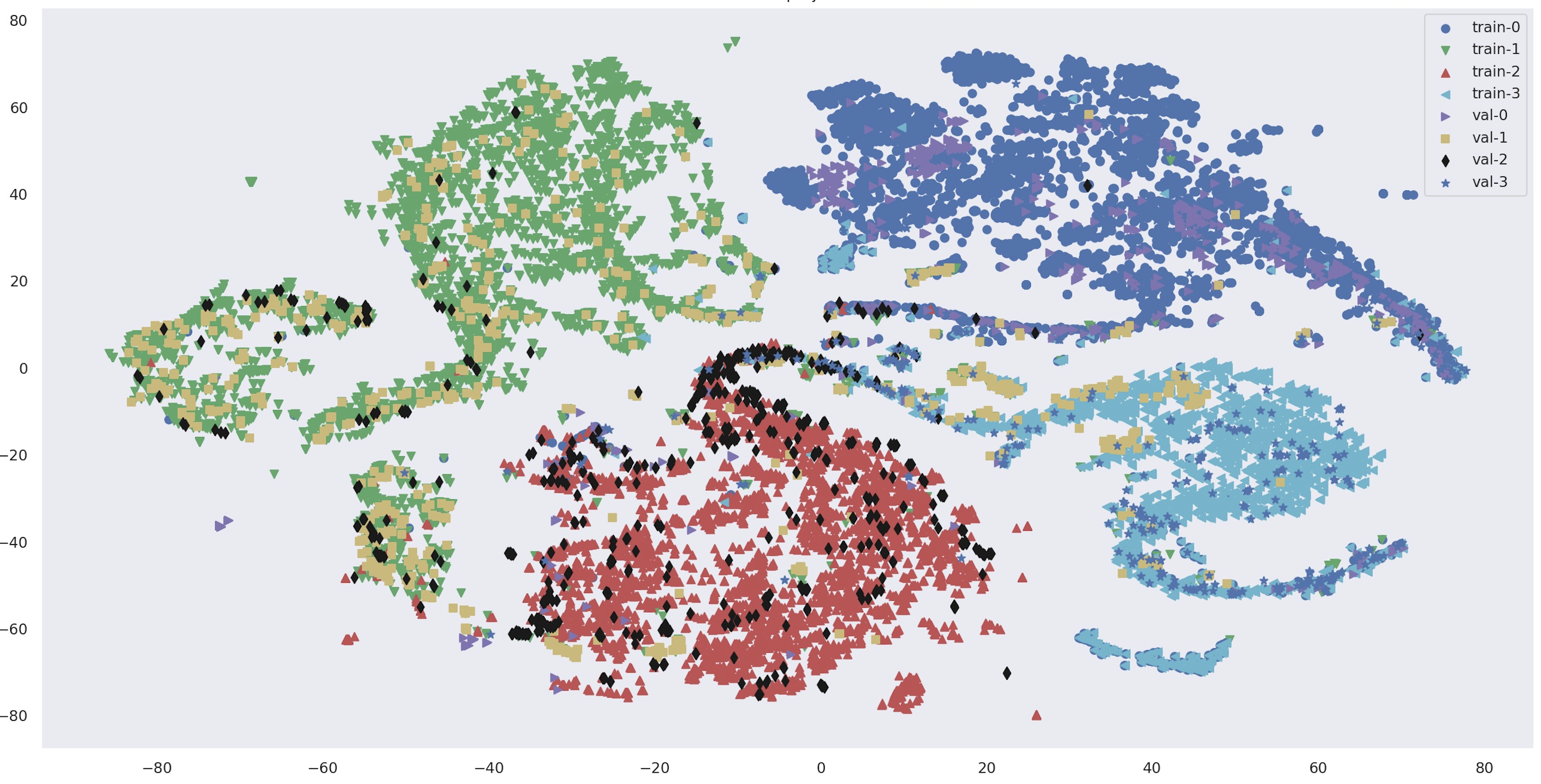}}
\caption{Data projection of validation set over a training set. Blue represents "Control" (train0/val0), and Green: Alcohol (train1/val1). Red: Drugs (train2/val2) and Light-blue: Sleepiness (train3/val3).}
\label{fig:tsne_val}
\end{figure}

Figure \ref{fig:tsne_val}, show the projection on the train and validation datasets. This image is relevant to understanding how the different classes are projected before training and identifying outliers.

\begin{figure}[]
\centerline
{\includegraphics[scale=0.20]{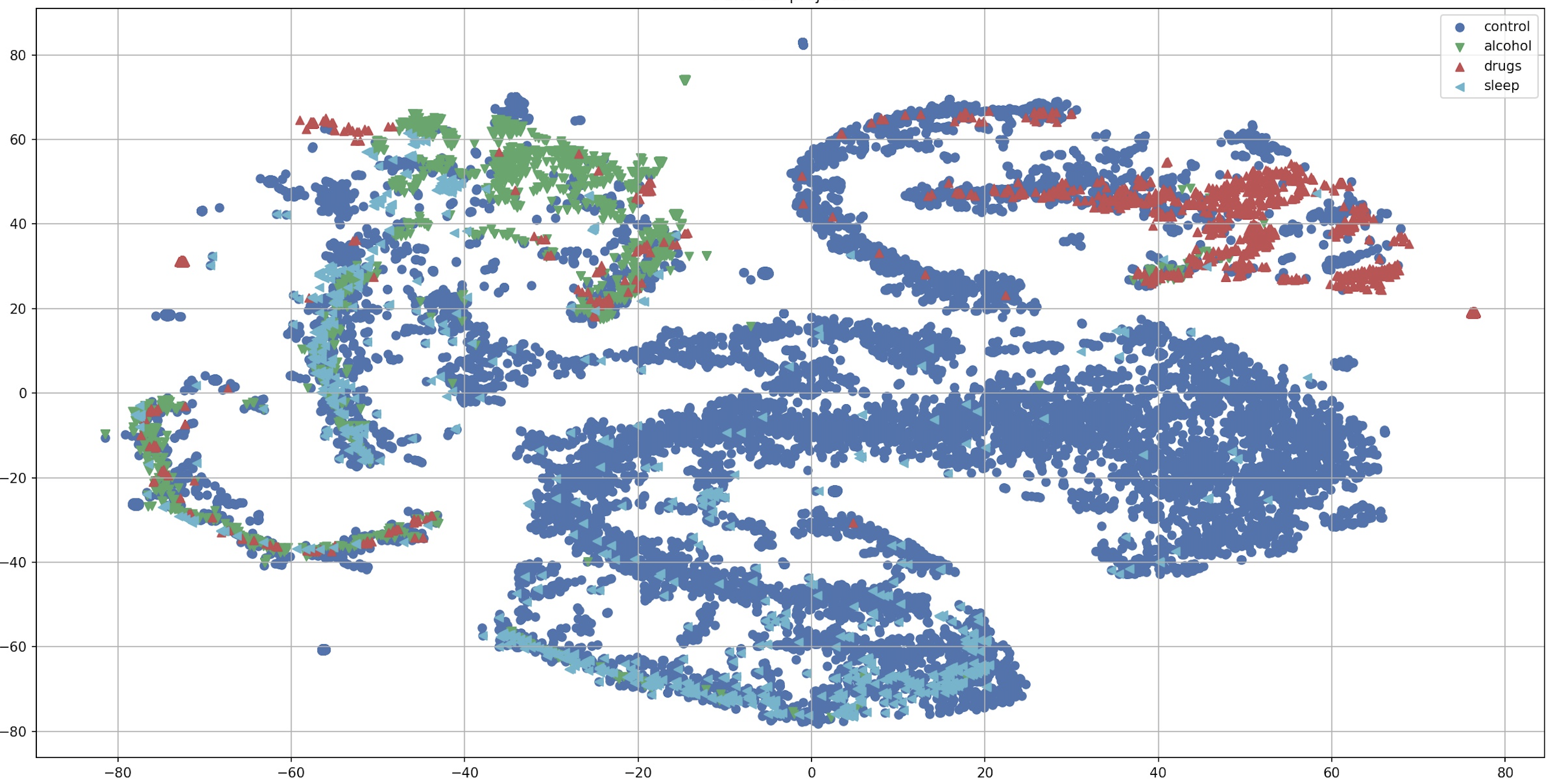}}
\caption{Data projection on the test set for the best model trained from scratch. Blues are Control subjects. Green represents alcohol, and Red represents drugs and light-blue sleepiness.}
\label{fig:tsne_test}
\end{figure}

Figure \ref{fig:tsne_test}, show the projection on the test dataset after classification. We can observe how the alcohol and drug classes are separated and the important overlapping of sleepiness examples. This projection adequately represents our results: sleepiness is identified as a more challenging class. It is essential to point out that all the subjects are volunteers then the control subject were labelled "control" according to their status. If we check the light blue (sleepiness), many of them are overlapped with control. That indicates that the subject did not sleep well (quantity of hours per day, on average 8 hours); however, he/she was self-declared as a control subject. This dataset represents very well the real problem in real conditions.

\begin{figure}[H]
\centering
\includegraphics[scale=0.50]{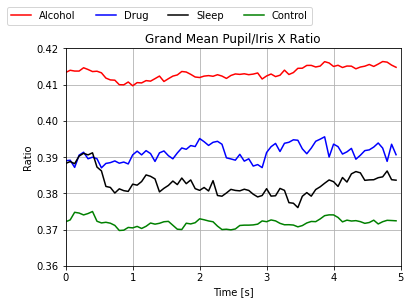}
\caption{\label{fig:ratio_pupil} Behavioural estimation of the iris images per subject in the presence of alcohol, drugs, and sleepiness.}
\end{figure}

In short, if we analyse all the sequences captured for all the subjects using a Grand mean algorithm, the behavioural curve can be estimated as shown in Figure \ref{fig:ratio_pupil}. These curves help to show the differences among each factor for control, alcohol, drugs and sleepiness. The alcohol in red appears as a factor easier to detect in comparison to sleepiness  (black) and drugs (blue).

\section{Conclusion}
\label{sec:conclusion}

This research offers new insights and opportunities into industrial biometric applications. In this work, we show that it is possible to capture the changes in the iris with a regular NIR sensor and estimate the alertness state of each person related to CNS. This novel estimation allows us to prevent accidents and save lives.

Sleepiness detection is identified as the most challenging FFD factor to be detected. Sleepiness is usually correlated with alcohol and drug consumption, as shown in the t-SNE. Subjects under pure sleepiness influences should be captured as included in the database in order to improve this proposal.

This study showed essential differences in pupil and iris behaviour under different conditions that may affect the behaviour of the CNS. The differences between the groups were quite remarkable, especially between the control cases and the subjects in the alcohol and drug data subsets. On the other hand, the behaviours of the drowsy or sleep-deprived subjects were closer to the control group. 

It is important to highlight that our results show that it is possible to detect, on average, 9 out of 10 subjects in Unfit conditions without altering the regular operation of an industry since the system has a specificity of more than 95\% for control (fit) subjects. Furthermore, our database replicates the expected behaviour of this type of operation, in which 10\% to 15\% of the worker present these conditions.

Our proposed system could be operated as a concurrent observation of the subject, e.g. medics, commercial pilots, air-forces, truck drivers, miner companies and any risk activity, without disturbing his primary duties. Thus, over time with the analysis over multiple time windows, a further improvement of the classification results can be achieved.

In future work, we will continue capturing sleepiness images to improve the results in this scenario and also evaluate Recurrent Convolutional Neural (R-CNN) networks and Long Short-Term Memory (LSTM) to include the time information in the classification task. As we demonstrate in these experiments, using 3 or 5 sequential frames without time correlation does not improve the results. 
\vspace{-0.5cm}
\ifCLASSOPTIONcompsoc
  \section*{Acknowledgments}
\else
  \section*{Acknowledgment}
\fi

This work is fully supported by the Agencia Nacional de Investigacion y Desarrollo (ANID) throught FONDEF IDEA N ID19I10118 leading by Juan Tapia Farias - DIMEC-UChile. And the German Federal Ministry of Education and Research and the Hessen State Ministry for Higher Education, Research and the Arts within their joint support of the National Research Center for Applied Cybersecurity ATHENE. Further thanks to all the volunteers that participated in this research.

\bibliographystyle{IEEEtran}
\bibliography{sample.bib}

\begin{IEEEbiography}[{\includegraphics[width=1in,height=1.25in,clip,keepaspectratio]{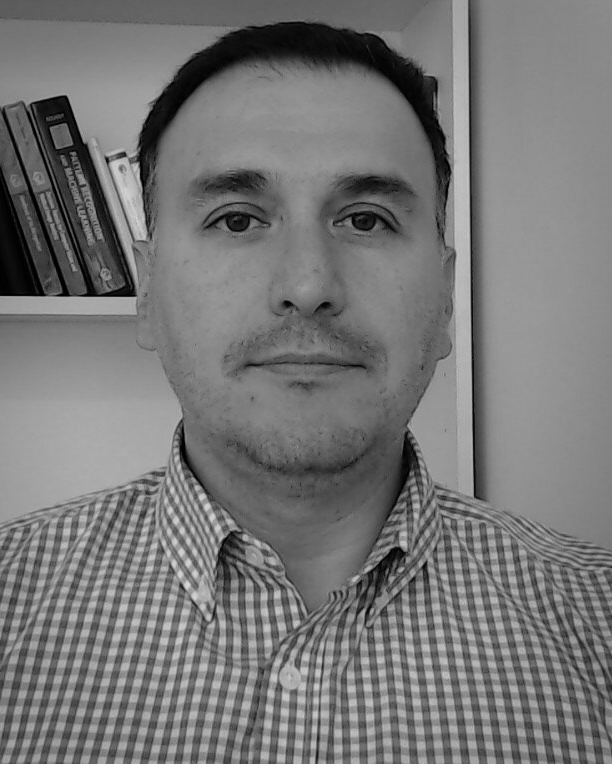}}]{Juan Tapia} received a P.E. degree in Electronics Engineering from Universidad Mayor in 2004, a M.S. in Electrical Engineering from Universidad de Chile in 2012, and a PhD from the Department of Electrical Engineering, Universidad de Chile in 2016. In addition, he spent one year of internship at the University of Notre Dame. In 2016, he received the award for best PhD thesis. From 2016 to 2017, he was an Assistant Professor at Universidad Andres Bello. From 2018 to 2020, he was the R\&D Director for the area of Electricity and Electronics at Universidad Tecnologica de Chile - INACAP. He is currently a Senior Researcher at Hochschule Darmstadt(HDA), and R\&D Director of TOC Biometrics. His main research interests include pattern recognition and deep learning applied to iris biometrics, morphing, feature fusion, and feature selection. 
\end{IEEEbiography}
\vspace{-7pt}

\begin{IEEEbiography}[{\includegraphics[width=1.10in,height=1.25in,clip,keepaspectratio]{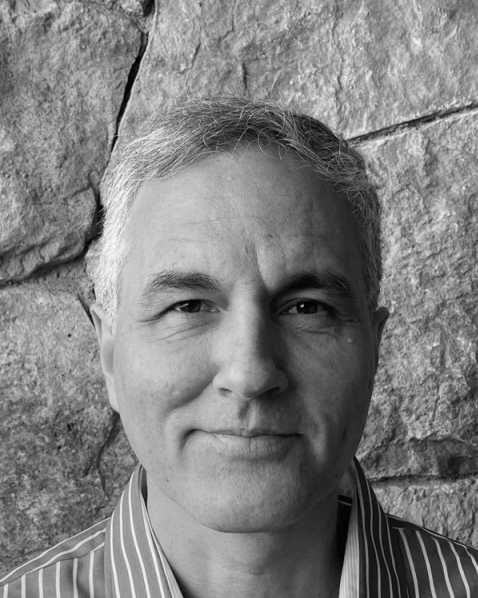}}]{Enrique Lopez Droguett} Droguett is a Professor in the Civil \& Environmental Engineering Department and the Garrick Institute for the Risk Sciences at the University of California, Los Angeles (UCLA), USA, and Associate Editor for both the Journal of Risk and Reliability and the International Journal of Reliability and Safety. He also serves on the Board of Directors of the International Association for Probabilistic Safety Assessment and Management (IAPSAM). Prof. López Droguett conducts research on Bayesian inference, and artificial intelligence-supported digital twins and prognostics and health management based on physics-informed deep learning for reliability, risk, and safety assessment of structural and mechanical systems. His most recent focus has been on quantum computing and quantum machine learning for developing solutions for risk and reliability quantification and energy efficiency of complex systems, particularly those involved in renewable energy production. He has led many major studies on these topics for a broad range of industries, including oil and gas, nuclear energy, defence, civil aviation, mining, renewable and hydro energy production and distribution networks. L{ó}pez Droguett has authored more than 250 papers in archival journals and conference proceedings.
\end{IEEEbiography}

\begin{IEEEbiography}[{\includegraphics[width=1in,height=1.25in,clip,keepaspectratio]{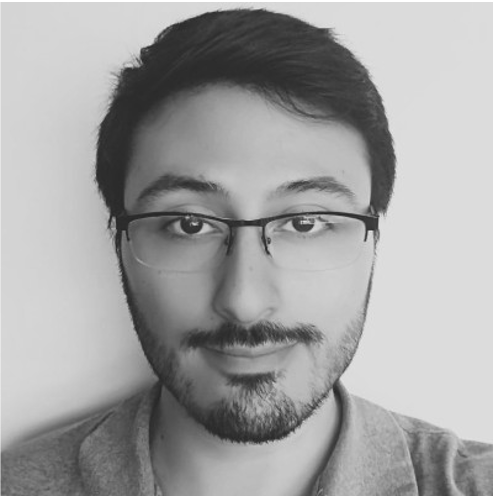}}]{Andres Valenzuela} received a B.S. in Computer Engineering from Universidad Andres Bello, Faculty of Engineering in Santiago, Chile in 2020. His main interests include computer vision, pattern recognition and Deep learning applied to semantic segmentation problems, focusing on NIR and RGB eyes images.
\end{IEEEbiography}

\begin{IEEEbiography}[{\includegraphics[width=1in,height=1.25in,clip,keepaspectratio]{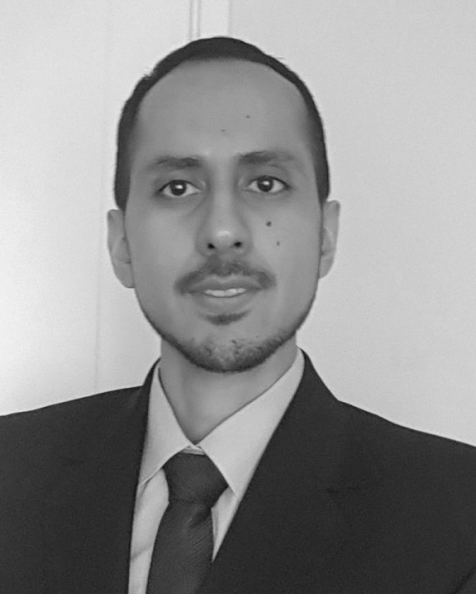}}]
{DANIEL P. BENALCAZAR} (M ‘09) was born in Quito, Ecuador in 1987. He obtained a B.S. in Electronics and Control Engineering from Escuela Politecnica Nacional, Quito, Ecuador in 2012. He received the M.S. in Electrical Engineering from The University of Queensland, Brisbane, Australia, in 2014, with a minor in Biomedical Engineering. He obtained a PhD in Electrical Engineering from Universidad de Chile, Santiago, Chile in 2020. From 2015 to 2016, he worked as a Professor at the Central University of Ecuador. Ever since he has participated in various research projects in biomedical engineering and biometrics. Mr. Benalcazar currently works as a researcher at TOC Biometrics Chile.
\end{IEEEbiography}

\begin{IEEEbiography}[{\includegraphics[width=1in,height=1.25in,clip,keepaspectratio]{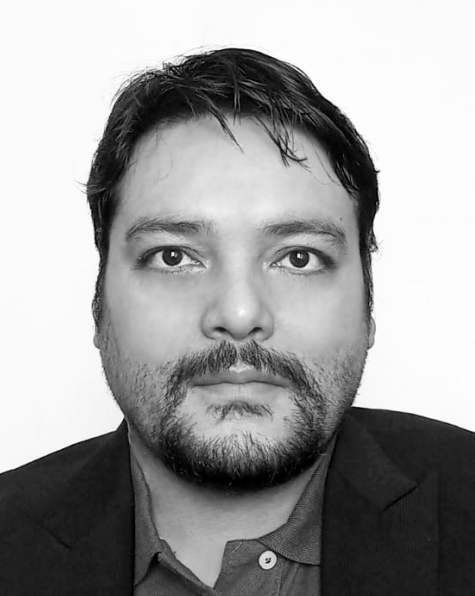}}]{Leonardo Causa} received the P.E. degree in electrical engineering from the Universidad de Chile in 2012, and the M.S. degree in biomedical engineering (BME) from the Universidad de Chile in 2012, he is also finishing the PhD degree in Electrical Engineering and Medical Informatics by cotutelage from U. de Chile and Universite Claude Bernard Lyon 1. His research interests include sleep pattern recognition, signal and image processing, neuro-fuzzy systems applied to the classification of physiological data, and machine and deep learning. He was engaged in research on automated sleep-pattern detection and respiratory signal analysis, fitness for duty, human fatigue, drowsiness, alertness and performance.
\end{IEEEbiography} 
\vspace{-10pt}

\begin{IEEEbiography}[{\includegraphics[width=1in,height=1.25in,clip,keepaspectratio]{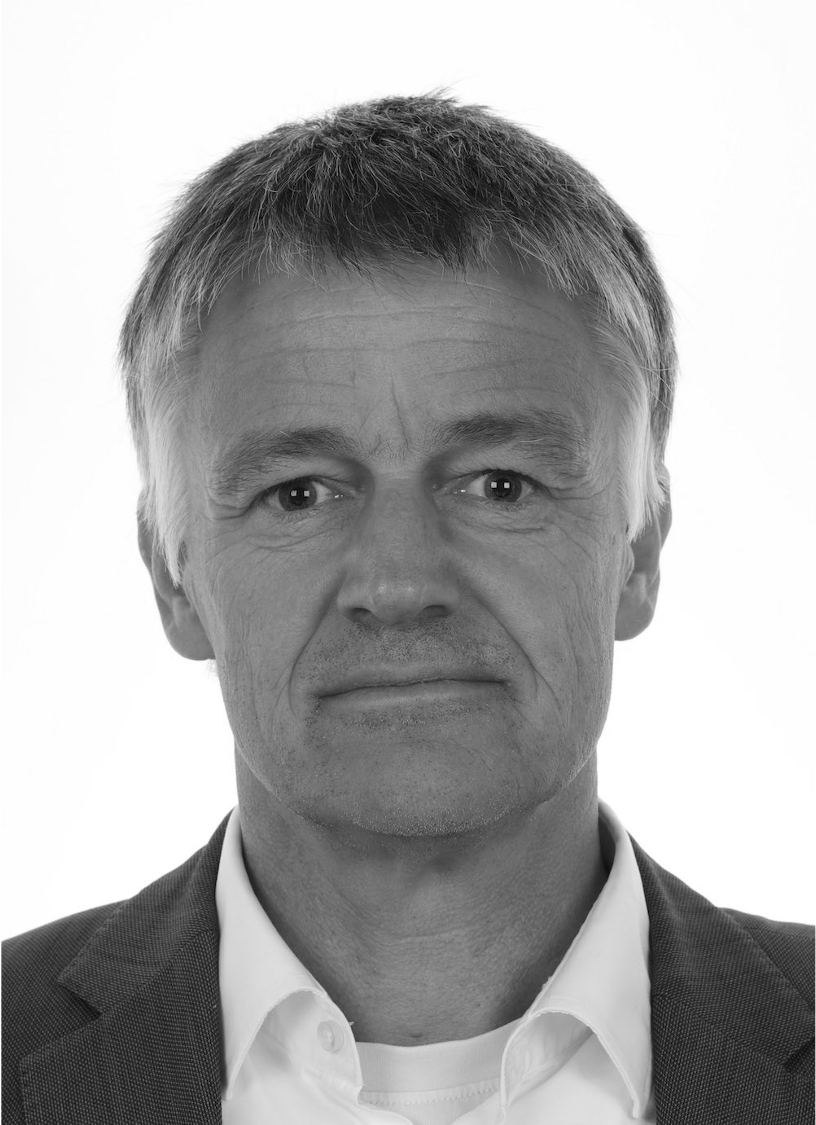}}]{Christoph Busch} is member of the Department of Information Security and Communication Technology (IIK) at the Norwegian University of Science and Technology (NTNU), Norway. He holds a joint appointment with the computer science faculty at Hochschule Darmstadt (HDA), Germany. Further, he has lectured the course Biometric Systems at Denmark’s DTU since 2007. On behalf of the German BSI, he has been the coordinator for the project series BioIS, BioFace, BioFinger, BioKeyS Pilot-DB, KBEinweg and NFIQ2.0. In the European research program, he was the initiator of the Integrated Project 3D-Face, FIDELITY and iMARS. Further, he was/is a partner in the projects TURBINE, BEST Network, ORIGINS, INGRESS, PIDaaS, SOTAMD, RESPECT and TReSPAsS. He is also a principal investigator in the German National Research Center for Applied Cybersecurity (ATHENE). Moreover, Christoph Busch is a co-founder and member of the board of the European Association for Biometrics (www.eab.org) which was established in 2011 and assembled in the meantime more than 200 institutional members. Christoph co-authored more than 500 technical papers and has been a speaker at international conferences. He is a member of the editorial board of the IET journal. 
\end{IEEEbiography}

\end{document}